\documentclass[letterpaper, 10 pt, conference]{ieeeconf}  

\IEEEoverridecommandlockouts                              

\usepackage{cite}
\usepackage{amsmath,amssymb,amsfonts}
\usepackage{algorithmic}
\usepackage{graphicx}
\def\BibTeX{{\rm B\kern-.05em{\sc i\kern-.025em b}\kern-.08em
    T\kern-.1667em\lower.7ex\hbox{E}\kern-.125emX}}
\usepackage{textcomp}
\usepackage{xcolor}
\usepackage{booktabs}
\usepackage{url}
\usepackage{float} 
\usepackage{hyperref}
\usepackage{siunitx}

\title{\LARGE \bf
UltraTac: Integrated Ultrasound-Augmented Visuotactile Sensor for Enhanced Robotic Perception
}

\author{Junhao Gong$^{1*}$, Kit-Wa Sou$^{1*}$, Shoujie Li$^{1\dagger}$, Changqing Guo$^{1}$, Yan Huang$^{1}$, \\Chuqiao Lyu$^{1}$, Ziwu Song$^{1}$, Wenbo Ding$^{1,2\dagger}$%
\thanks{*These authors contributed equally to this work.}%
\thanks{This work was supported by National Key R\&D Program of China (No.2024YFB3816000), Shenzhen Key Laboratory of Ubiquitous Data Enabling (No. ZDSYS20220527171406015), Guangdong Innovative and Entrepreneurial Research Team Program (2021ZT09L197), and Tsinghua Shenzhen International Graduate School-Shenzhen Pengrui Young Faculty Program of Shenzhen Pengrui Foundation (No. SZPR2023005), and Meituan Academy of Robotics Shenzhen.}%
\thanks{$^{\dagger}$Corresponding author: Shoujie Li (lsj20@mails.tsinghua.edu.cn), Wenbo Ding (ding.wenbo@sz.tsinghua.edu.cn)}%
\thanks{$^{1}$Shenzhen Ubiquitous Data Enabling Key Lab, Shenzhen International Graduate School, Tsinghua University, Shenzhen 518055, China.}%
\thanks{$^{2}$RISC-V International Open Source Laboratory, Shenzhen 518055, China.}%
\thanks{This paper has supplementary downloadable material available at: 
\href{ultratac.junhaogong.top}{ultratac.junhaogong.top}}%
}

\begin{document}

\maketitle
\thispagestyle{empty}
\pagestyle{empty}

\begin{abstract}

Visuotactile sensors provide high-resolution tactile information but are incapable of perceiving the material features of objects. We present UltraTac, an integrated sensor that combines visuotactile imaging with ultrasound sensing through a coaxial optoacoustic architecture. The design shares structural components and achieves consistent sensing regions for both modalities. Additionally, we incorporate acoustic matching into the traditional visuotactile sensor structure, enabling the integration of the ultrasound sensing modality without compromising visuotactile performance. Through tactile feedback, we can dynamically adjust the operating state of the ultrasound module to achieve more flexible functional coordination. Systematic experiments demonstrate three key capabilities: proximity sensing in the 3–8 cm range (R² = 0.99), material classification (average accuracy: 99.20\%), and texture-material dual-mode object recognition achieves 92.11\% accuracy on a 15-class task. Finally, we integrate the sensor into a robotic manipulation system to concurrently detect container surface patterns and internal content, which verifies its promising potential for advanced human-machine interaction and precise robotic manipulation.
\end{abstract}


\section{INTRODUCTION}

Visuotactile sensors (e.g., GelSight\cite{gelsight}, DIGIT\cite{digit}, GelSlim\cite{gelslim}) capture high-resolution optical images of surface deformations to reveal texture, contact location, and force distribution. Nonetheless, they acquire data only during contact and cannot detect material properties\cite{kappassov2015, wang2022}. This limitation challenges robotic systems requiring proximity sensing\cite{li2020}. Moreover, the inability to ascertain material characteristics, including internal cavities and content identification, compromises safety and efficiency in dynamic environments, particularly when handling delicate or unfamiliar objects\cite{shimonomura2019}.

\begin{figure}[t]
    \centering
    \includegraphics[width=\columnwidth]{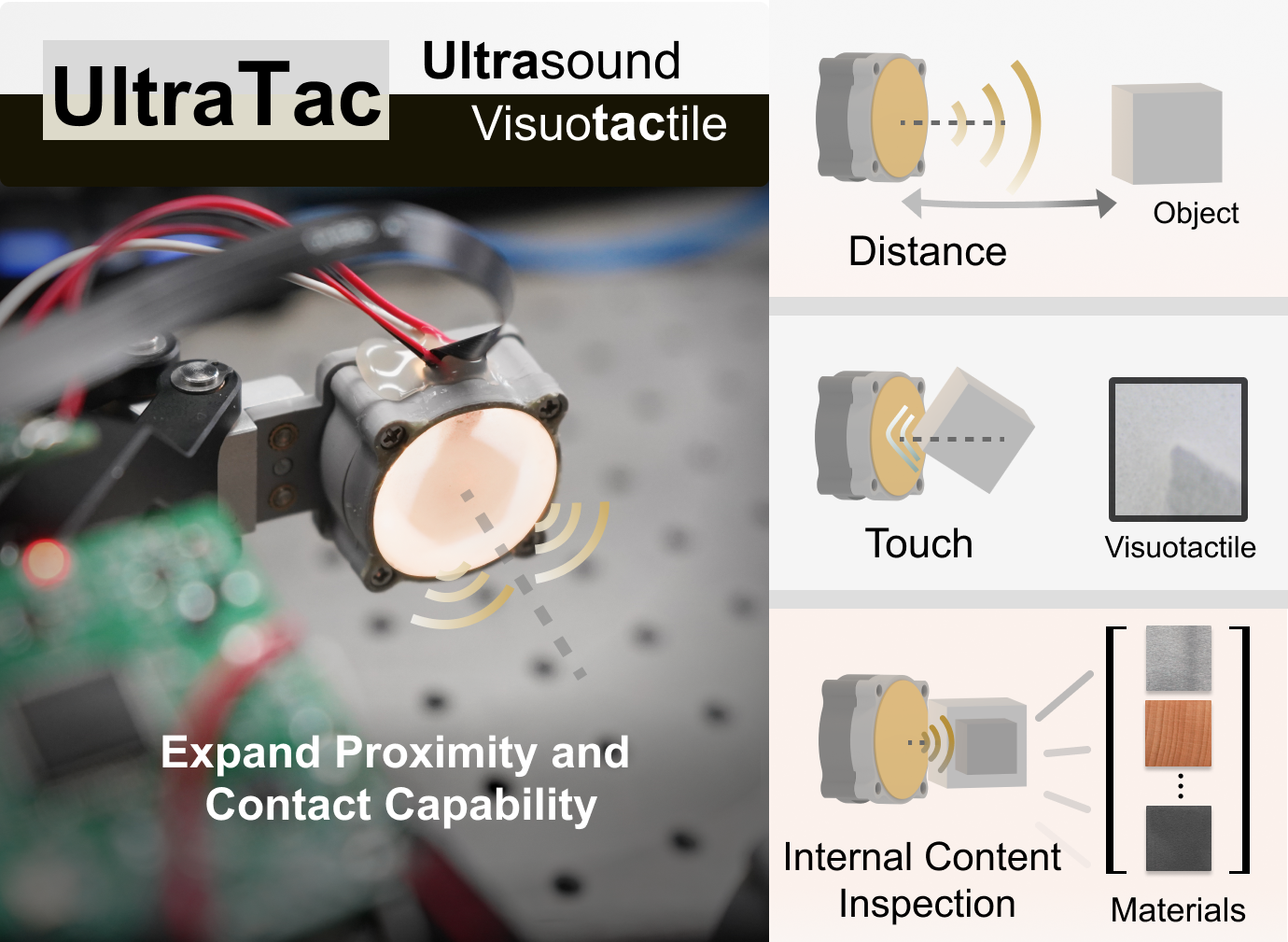} 
    \caption{Overview of UltraTac: an integrated sensor combining visuotactile and ultrasound sensing.}
    \label{fig:FigOverview}
\end{figure}

To address these limitations, the integration of multiple sensing modalities is explored in various contexts. The work presented in \cite{WACV2021, RAL2022} demonstrates a novel approach to proximity sensing by employing selective membrane transparency and integrated internal light sources, thereby enabling the accurate detection of the distance between objects and sensors prior to physical contact. However, this approach encounters challenges in dark environments or those with complex lighting conditions. Furthermore, the integration of both sensing modalities using a single camera with time-division multiplexing exacerbates the complexities associated with light source control and data acquisition. Song et al.\cite{NanoEnergy2022} employ a flexible triboelectric sensor to achieve simultaneous material and texture recognition, while Lee et al.\cite{AS2021} employ hybrid triboelectric, piezoelectric, and piezoresistive operating mechanisms to simultaneously distinguish between materials and textures. However, both approaches face a common challenge: extracting two distinct modalities from a unified signal may lead to crosstalk, where interference between the modalities degrades signal fidelity and classification performance.

To overcome the limitations of traditional visuotactile sensors, ultrasound sensing adds value by measuring distances without contact and analyzing materials beneath the surface. Operating in a proximity sensing mode, ultrasound measures distances before contact is made and provides material features by analyzing reflected acoustic waves \cite{material detection2023}. Recent studies show that ultrasound sensors detect objects at distances of several centimeters with high precision \cite{cho2017}, making them valuable for proximity sensing. Additionally, when in contact with objects, ultrasound penetrates surfaces to reveal internal structures and material properties that remain invisible to optical methods \cite{Pavlin2017}.

In this paper, we introduce UltraTac, an innovative integrated sensor that effectively tackles these challenges by seamlessly merging visuotactile imaging with ultrasound sensing via a coaxial optoacoustic design (Fig.~\ref{fig:FigOverview}). Our study presents three major contributions:

\begin{itemize}
    \item \textbf{Novel integrated sensor architecture:} We develop a compact sensor that unifies visuotactile imaging with ultrasound through a coaxial design. By using a translucent, ultra-thin elastomer and an annular lead zirconate titanate (PZT) transducer with an engineered impedance matching stack, we achieve optimal acoustic coupling while preserving optical clarity for simultaneous tactile and ultrasound sensing.
    
    \item \textbf{Enhanced sensing capabilities via ultrasound augmentation:} Our system provides two capabilities unavailable to conventional tactile sensors: proximity sensing and contact-based material classification. These modes can be dynamically interchanged based on tactile feedback, enabling adaptive perception across diverse interaction scenarios.
    
    \item \textbf{Extensive experimental validation and application demonstrations:} Through systematic experiments, we validate the sensor's effectiveness in proximity detection, material classification, and dual-modal object recognition tasks, demonstrating its potential for robotic manipulation, content inspection, and other applications requiring object assessment.
\end{itemize}

\section{Related Work}
In recent years, multimodal visuotactile sensors have significantly advanced robotic perception. Comprehensive reviews by Luo et al. \cite{luo2017} and Dahiya et al. \cite{dahiya2010} highlight the benefits of multimodal approaches in addressing the limitations of individual sensing technologies. 

Proximity sensing enables early anticipation of interactions, enhancing performance in complex manipulation tasks. CompdVision integrates a compound-eye imaging system that employs far-focus stereo units for external depth estimation and near-focus units for tactile deformation tracking, enabling accurate pre-contact object localization\cite{CompdVision}. Similarly, the multimodal fingertip sensor by SaLoutos et al. combines embedded pressure sensors with time-of-flight proximity modules to measure contact forces and detect approaching objects\cite{ICRA2023}. Moreover, Li et al.’s M\textsuperscript{3}Tac extends sensor capability by fusing visible, near-infrared, and mid-infrared imaging, thereby achieving high-resolution proximity sensing, precise three-dimensional reconstruction, and temperature measurement\cite{TRO2024}.

For material classification specifically, researchers demonstrate that combining different sensing modalities can improve discrimination between visually similar materials \cite{xu2019, yuan2017}. Song et al. \cite{song2023} combine tactile and temperature sensing for enhanced material recognition, while Sou et al. \cite{sou2024} integrate mechanoluminescence with tactile sensing for event-trigger perception. Abderrahmane et al. \cite{abderrahmane2018} show how multiple sensing modalities improve object recognition even for previously unseen items.

Non-destructive testing and subsurface analysis are critical areas where traditional tactile sensing falls short. Current ultrasound methods for internal inspection \cite{mao2024} show promise but typically require specialized equipment separate from tactile systems. Capturing both surface properties and internal characteristics within a single compact sensor represents an advancement for applications ranging from quality control to medical diagnostics \cite{Davis2020}.

Nonetheless, seamlessly integrating tactile and ultrasound sensing into a unified sensor architecture remains an open challenge. Prior attempts to combine these modalities often yield bulky systems with separate sensing elements, limiting their practical application in robotic end-effectors \cite{kerr2018}. Moreover, maintaining optical clarity for visuotactile sensing alongside efficient acoustic coupling for ultrasound necessitates innovative solutions.

\begin{figure}[ht]
    \centering
    \includegraphics[width=1.\columnwidth]{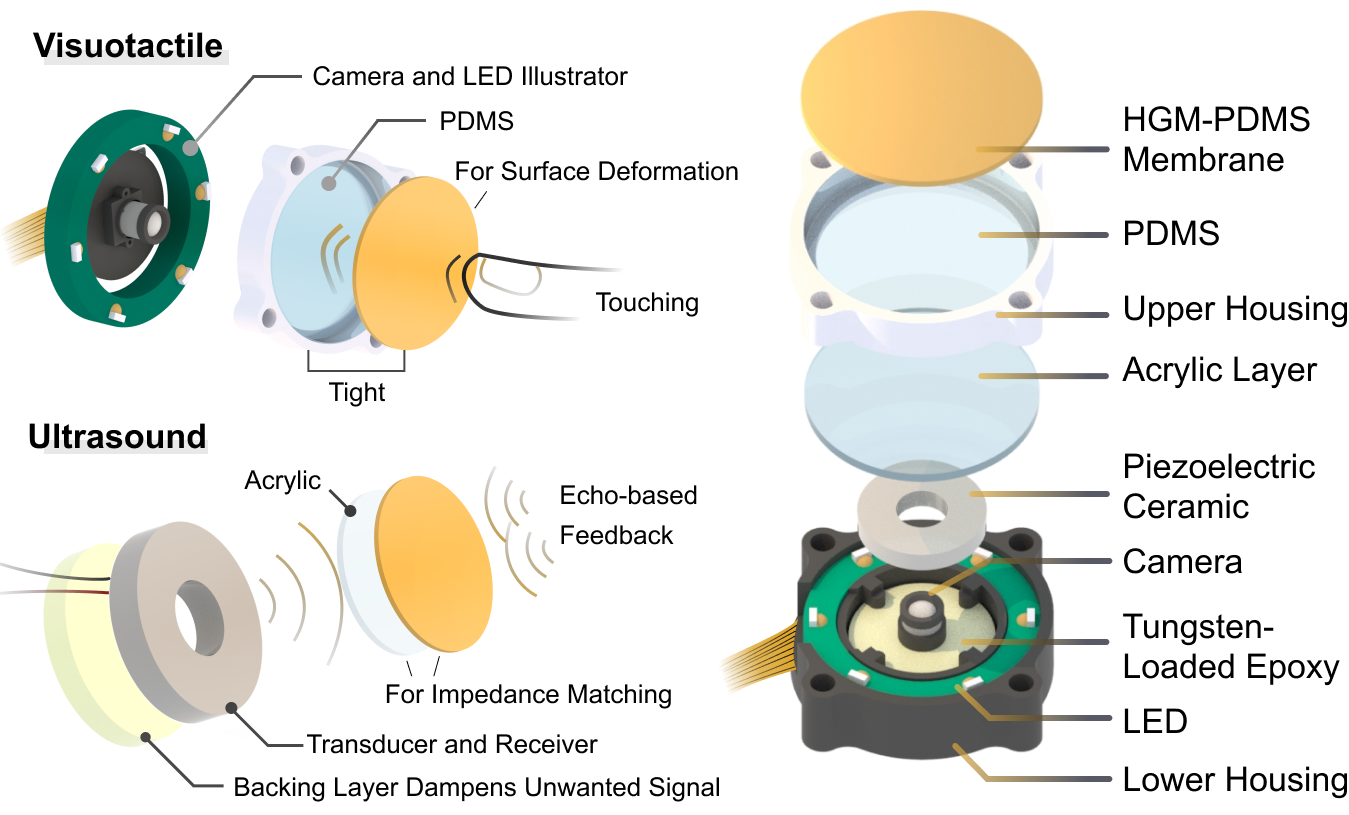} 
    \caption{Structure illustration showing the visuotactile and ultrasound modules with exploded component view.}
    \label{fig:Figure2}
\end{figure}
\vspace{-2mm}

\section{Design and Implementation}

Traditional visuotactile sensors integrate a camera, acrylic substrate, flexible deformation layer, and light-blocking membrane for optical imaging and tactile detection. However, this structure impedes ultrasound propagation due to acoustic impedance mismatches between layers. Our design challenge was to create a structure to enable effective ultrasound transmission while preserving visuotactile capabilities.

\subsection{Design Principles} 

UltraTac delivers dual-modal perception by integrating surface texture information from camera imaging through an elastomer membrane with material properties detected via ultrasound, as shown in Fig.~\ref{fig:Figure2}. We optimize the sensor's mechanics, materials, electronic systems, and algorithms to achieve effective dual-modal sensing capabilities.

Mechanically, UltraTac is built around a coaxial optoacoustic structure, ensuring that the camera and the ring-shaped piezoelectric ceramic (PZT) share the same central axis. This layout preserves the optical path for visuotactile sensing while simultaneously providing an unobstructed route for ultrasound transmission and reception. This structural integration ensures consistent sensing regions for both modalities, enhancing the spatial alignment of visuotactile and ultrasound sensing within the sensor. 

From a materials perspective, we introduce several unconventional materials in the visuotactile domain to achieve effective acoustic impedance matching. In parallel, the electronic systems have been designed as a highly integrated processing unit that supports both camera and ultrasound functionalities. Furthermore, from an algorithmic standpoint, the system employs a touch-triggered dual-pathway approach that dynamically adapts sensor functionality; when no contact is detected, ultrasound performs proximity sensing, whereas upon contact it switches to material classification while visual data simultaneously processes texture recognition.

\vspace{-2mm}
\begin{table}[htbp]
\centering
\caption{Acoustic impedance of different materials}
\label{tab:acoustic_impedance}
\begin{tabular}{cc}
\toprule
\textbf{Materials} & \textbf{Acoustic Impedance (MRayls)} \\
\midrule
Tungsten & 100\cite{MDPI} \\
Epoxy & 3-4\cite{MDPI} \\
PZT & 25-35\cite{MDPI} \\
Acrylic & 3.2\cite{Fundamentals of acoustics} \\
PDMS & 1.1\cite{PDMS} \\
Hollow Glass Microsphere & 0.2\cite{HGM} \\
Air & 0.000415\cite{Fundamentals of acoustics} \\
\bottomrule
\end{tabular}
\label{Acoustic impedance}
\end{table}
\vspace{-3mm}

\subsection{Acoustic Matching}
In this work, we employ acoustic impedance matching principles based on two key formulas. The first is the single-layer matching formula\cite{Fundamentals of acoustics}:
\begin{equation}
    Z_m = \sqrt{Z_1 \cdot Z_2},
    \label{eq:impedance_matching}
\end{equation}
where \(Z_1\) and \(Z_2\) are the acoustic impedance of the two media and \(Z_m\) is the optimal impedance of the matching layer. The second formula is the quarter-wavelength thickness condition:
\begin{equation}
    d = \frac{\lambda_m}{4} = \frac{c_m}{4f},
    \label{eq:quarter_wavelength}
\end{equation}
where \(d\) is the matching layer thickness, \(\lambda_m\) is the wavelength in the layer, \(c_m\) is the speed of sound in the layer, and \(f\) is the operating frequency. These formulas, together with the data in Table~\ref{Acoustic impedance}, will be used for acoustic matching.

From a materials standpoint, we aim to enhance acoustic transmission efficiency while preserving optical transparency. Typically, the light-blocking layer in sensors consists of flexible membranes infused with metal powders or dyes. However, as shown in Table~\ref{Acoustic impedance}, these membranes exhibit significantly higher acoustic impedance than air, leading to substantial reflection of ultrasound signals at the interface \cite{Fundamentals of acoustics}. To mitigate this loss, we incorporate hollow glass microspheres (HGM) into the dyed membrane, thereby reducing the impedance mismatch between the PDMS layer and air. HGM is widely employed to lower acoustic impedance, with its effectiveness dependent on the volume fraction and particle size \cite{HGM}. In our work, HGM is mixed with dyed PDMS at a 1:1 volume ratio to function both as a light-blocking layer for imaging and as an acoustic matching layer at the PDMS–air interface. The composite layer's optimal impedance was determined using the single-layer matching formula (Equation~\eqref{eq:impedance_matching}), ensuring enhanced transmission efficiency.

\begin{figure}[ht]
    \centering
    \includegraphics[width=1\columnwidth]{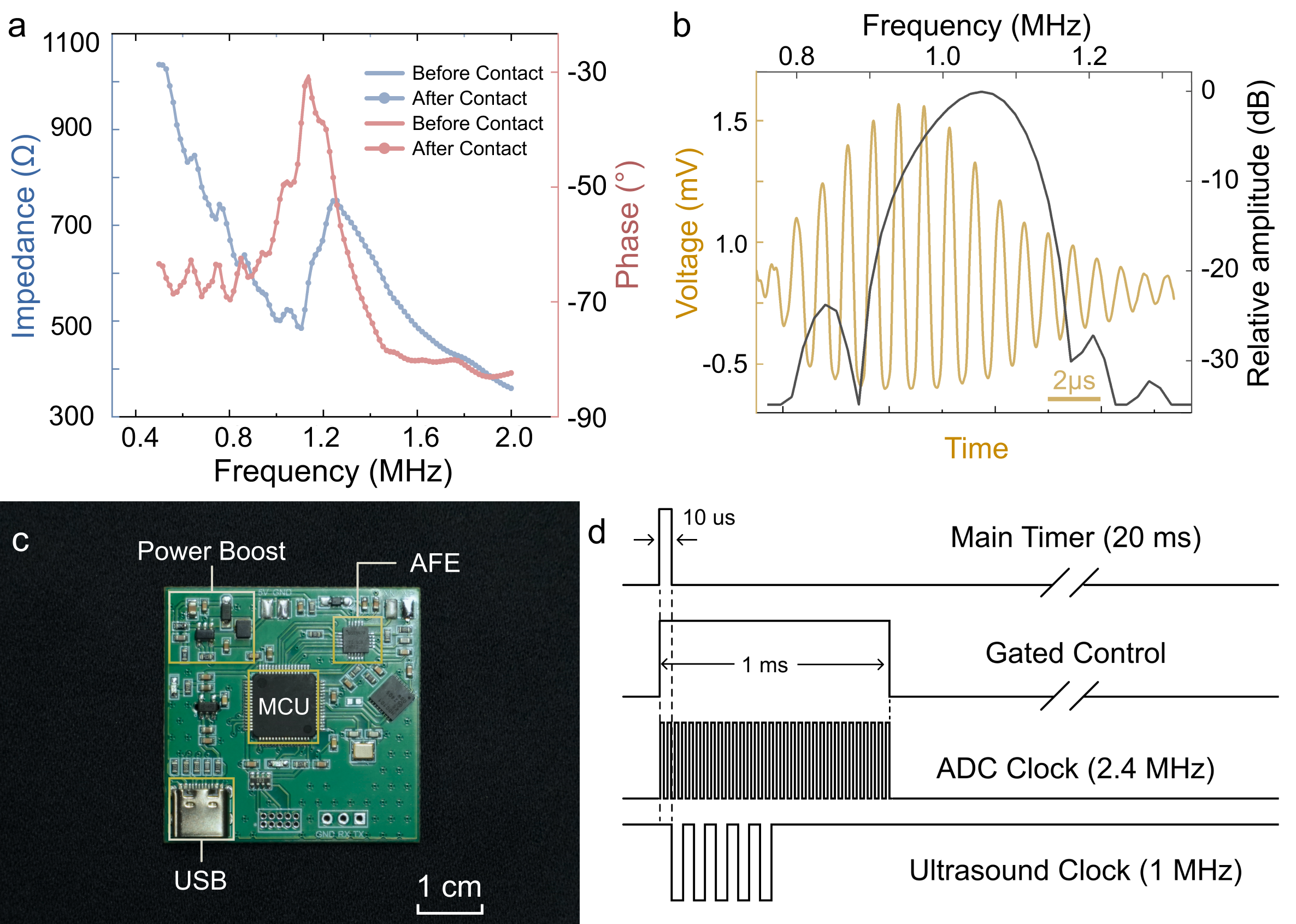} 
    \caption{Ultrasound transducer characteristics. (a) Impedance and phase response across frequency. (b) Time-domain signal pulse and frequency response. (c) The PCB of the ultrasound module. (d) Timing diagram of ultrasound module.}
    \label{fig:FigUltraBasic}
\end{figure}

Under the elastomer layers, an acrylic substrate is typically used to support the deformation of the elastomer. To satisfy the acoustic impedance matching conditions, rather than altering the material type, we adjust the acrylic thickness to 0.7 mm, calculated based on the propagation speed of 1 MHz ultrasound in acrylic and the quarter-wavelength formula(Equation~\eqref{eq:quarter_wavelength}). This optimized thickness allows the acrylic to function as a single-layer matching medium between the PZT and PDMS.

Additionally, an ultrasound sensor requires a backing layer to suppress undesired acoustic reflections. Based on Equation~\eqref{eq:impedance_matching}, tungsten-loaded epoxy is widely used in ultrasound applications as a backing-layer material because its acoustic impedance can be tuned by adjusting the ratio of tungsten powder to resin\cite{MDPI}. In our design, we employ tungsten-loaded epoxy prepared at a 3:2 volume ratio.

\begin{figure*}[ht]
    \centering
    \includegraphics[width=1.81\columnwidth]{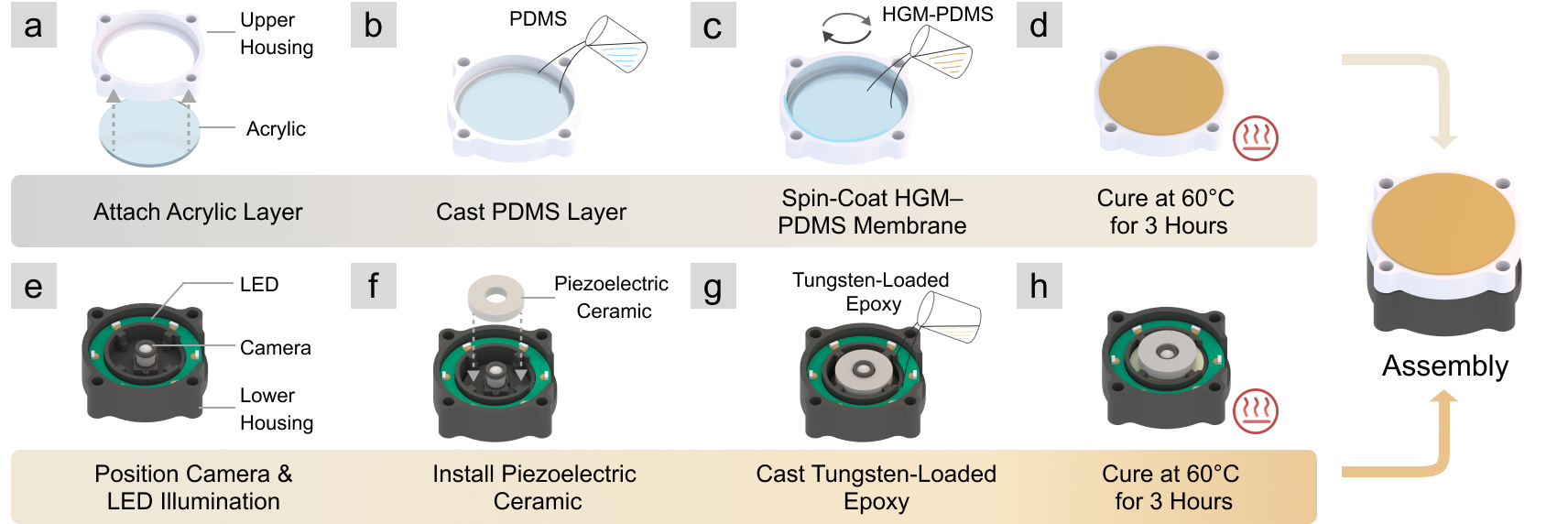} 
    \caption{Fabrication process of the sensor showing parallel assembly paths. (a-d) Upper housing with acrylic layer attachment, PDMS casting, HGM-PDMS spin-coating, and curing. (e-h) Lower housing with camera and LED positioning, piezoelectric ceramic installation, tungsten-epoxy casting, and final assembly of both components.}
    \label{fig:FigureFra}
\end{figure*}

We conduct an acoustic analysis of the entire sensor. Fig.~\ref{fig:FigUltraBasic}(a) presents the impedance and phase variations over the frequency range of 500 kHz to 2 MHz. The minimum impedance corresponds to the sensor's resonant frequency (1.05 MHz), while the maximum impedance is observed at the anti-resonant frequency (1.21 MHz). The marked curve represents the impedance response under applied pressure. The impedance characteristics remain largely unchanged before and after compression, indicating that the deformation of the flexible layer above the PZT does not affect the ultrasound functionality. This provides a fundamental physical basis for the integration of the tactile sensing modality. Fig.~\ref{fig:FigUltraBasic}(b) shows the echo signal obtained when the sensor is placed on a container filled with water. The yellow waveform represents the time-domain signal, while the black waveform corresponds to the frequency-domain signal obtained after performing a Fourier transform on the time-domain data. From the information presented in the figure, we observe that the sensor's central frequency is 1.05 MHz.

\subsection{Electronic Systems}

\textbf{Ultrasound Module:} To integrate ultrasound functionality into robotic systems, a compact and highly integrated ultrasound transmission and reception system is required. To address this need, a dedicated PCB (Fig.~\ref{fig:FigUltraBasic}(c)) was designed (4 cm × 4 cm) that requires only a single USB cable for both power and communication. The PCB, featuring a power boost circuit, generates excitation voltages up to 40 V for ultrasound signals and achieves a dynamic sensing refresh rate of approximately 50 Hz, meeting the real-time demands of human–machine interaction. The system architecture comprises a 144 MHz microcontroller unit (MCU) and a highly integrated analog front-end (AFE). The MCU dynamically adjusts both the number of transmitted ultrasound pulses and the gain in the reception chain, enabling flexible modulation of ultrasound functionality.

The timing diagram of the ultrasound acquisition system is shown in Fig.~\ref{fig:FigUltraBasic}(d). The MCU operates with a 20 ms cycle for both transmission and acquisition, utilizing a 2.4 MHz ADC to capture the echo envelope, with all triggers controlled by a hardware timer. During the idle period within the 20 ms cycle, the MCU transmits data and updates the operational status of the ultrasound system. Specifically, during the operation of the ultrasound distance measurement mode, the AFE is configured to emit five pulses per cycle to minimize the blind zone, while the reception chain gain is set to an amplification factor of 55.5 dB at 1 mV. Conversely, when the ultrasound content recognition mode is activated, the pulse count is increased to 20 in order to enhance recognition accuracy. The transition between these two operational modes is governed by haptic feedback; when the tactile modality detects contact between the sensor and an object, a contact trigger signal is transmitted to the MCU, thereby initiating the switch in ultrasound functionality.

\textbf{Optical Module:} The optical system consists of a micro-camera with a 1.88 mm focal length, positioned at the center of the annular piezoelectric ceramic. The camera lens, with a 6 mm diameter, is precisely integrated within the ceramic ring. A circular LED board is mounted around the outer perimeter of the piezoelectric ceramic, ensuring uniform illumination without interfering with ultrasound wave propagation. Experimental evaluations indicate that the imaging area closely approximates a 15 mm diameter circle, aligning with the dimensions of the ultrasound annular ring. This spatial consistency suggests a high degree of congruence between the optical and ultrasound sensing regions.

\subsection{Fabrication}
The fabrication process follows a systematic eight-step procedure as illustrated in Fig.~\ref{fig:FigureFra}, with parallel assembly paths for the upper and lower sensor components.

For the upper housing assembly (Fig.~\ref{fig:FigureFra} (a-d)), the process begins by attaching a thin acrylic layer (0.7 mm) to the bottom of the ring-shaped upper housing using cyanoacrylate adhesive 502. This acrylic layer serves dual functions as both an acoustic matching element and a deformation platform for the elastomer. A PDMS elastomer layer with a mass ratio of 30:1 (base to curing agent) is then cast into the housing structure, creating the base elastic layer for tactile sensing. Following this, an HGM–PDMS mixture (volume ratio 1:1) is applied using spin-coating at 3000 rpm for 30 s to form a thin, uniform flexible membrane over the PDMS layer. This specialized membrane functions as both the light-blocking layer for tactile imaging and acoustic matching layer for ultrasound transmission. The upper assembly is then heat-cured at 60 °C for three hours.

For the lower housing assembly (Fig.~\ref{fig:FigureFra} (e-h)), the camera module is precisely positioned at the center of the lower housing, with the LED illumination ring arranged around it to provide uniform lighting for tactile imaging while maintaining the central optical path. The annular PZT transducer is installed around the camera module, ensuring alignment with the central optical axis to maintain the coaxial design. A tungsten-loaded epoxy backing (volume ratio 3:2) is cast using a soft-tip syringe to fill the cavity behind the PZT transducer, creating the acoustic backing layer necessary for ultrasound performance. This lower assembly is also heat-cured at 60 °C for three hours. After fabricating and curing both the upper and lower assemblies, they are aligned and combined to form the coaxial sensor module.

\begin{figure}[ht]
    \centering
    \includegraphics[width=1.\columnwidth]{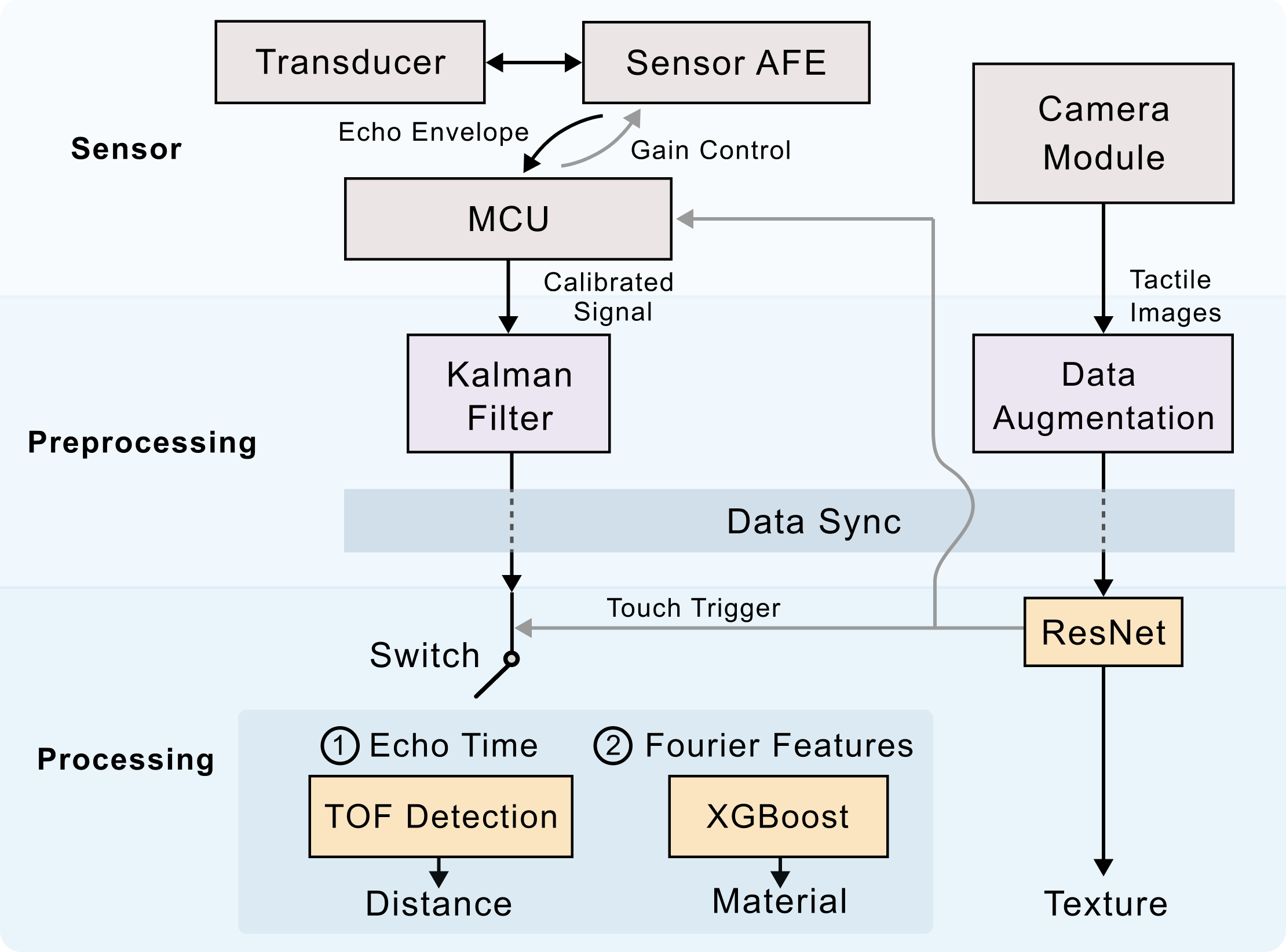} 
    \caption{Overview of the three-level ultrasound–visuotactile fusion pipeline.}
    \label{fig:FigWorkflow}
\end{figure}
\vspace{-2mm}

\subsection{System Pipeline}
The pipeline shown in Fig.~\ref{fig:FigWorkflow} is organized into three hierarchical levels. At the sensor level, the ultrasound path consists of a transducer connected to a sensor AFE, which extracts echo envelopes and communicates with the MCU for gain control. Parallel to this, the camera module captures tactile information.

At the preprocessing stage, ultrasound signals are Kalman filtered for noise reduction, while tactile images are augmented. A critical data sync mechanism aligns 50 Hz ultrasound data with 30 Hz visual data by pairing each ultrasound measurement with its nearest camera frame.

At the processing level, a dual-pathway approach is employed upon touch detection. A touch trigger signal from the camera path activates a switch that routes ultrasound data to either Time-of-Flight (ToF) detection for calculating object distance using echo time during non-contact phases, or XGBoost for material classification using Fourier features during contact. Simultaneously, a ResNet18 model processes tactile images for texture recognition\cite{resnet}.

\section{Experimental Evaluation}

To validate the sensor's capabilities, we conducted four systematic experiments: proximity detection, material classification, dual-modal classification, and internal content inspection.

\begin{figure}[ht]
    \centering
    \includegraphics[width=1\columnwidth]{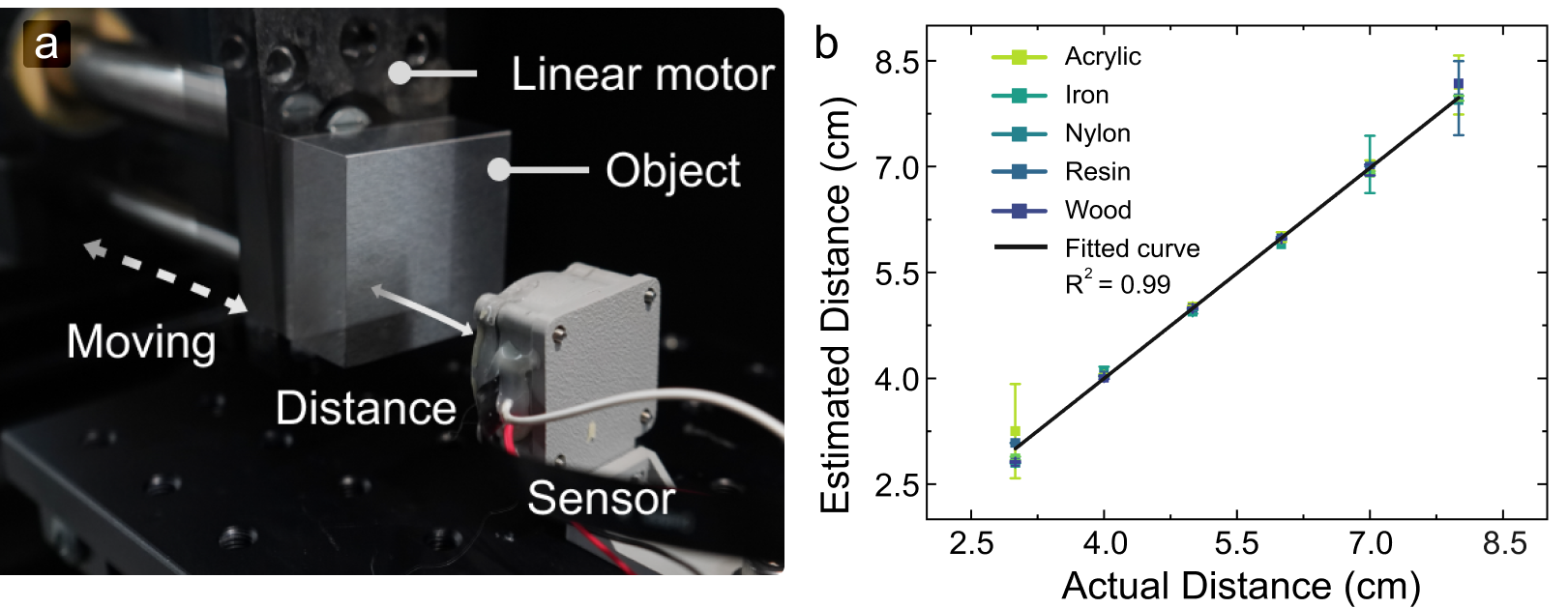} 
    \caption{Proximity detection experiment. (a) Experimental setup. (b) Comparison of estimated vs. actual distances.}
    \label{fig:prox_detect}
\end{figure}

    \subsection{Proximity Detection Experiment} 
    
    We exploit the propagation and reflection characteristics of ultrasound in air by measuring the ToF of the ultrasound signal in the time-domain. This approach allows the sensor to measure the distance to an object, providing valuable information before contact to assist with tasks like collision avoidance and trajectory planning. In our processing method, during sensor initialization, an echo signal (after noise removal) is recorded as a reference frame. Once the distance measurement function is activated, the current frame is subtracted point-by-point from the reference frame. The time corresponding to the voltage peak in the resulting difference frame is identified as the echo time, which is then used to calculate the predicted distance using the method described above.
    
    We evaluate the sensor's proximity sensing capabilities using the linear motor stage setup shown in Fig.~\ref{fig:prox_detect}(a), where objects of five different materials (acrylic, iron, nylon, resin, and wood) are mounted at distances ranging from 3.0 cm to 8.0 cm from the sensor.
    
    Fig.~\ref{fig:prox_detect}(b) demonstrates the system's measurement accuracy, with estimated distances strongly correlating with actual distances (R² = 0.99). The error bars indicate measurement consistency across all five materials, despite their distinct acoustic impedance. The linear relationship between estimated and actual distances remains robust throughout the tested range, with average estimation errors below ±0.5 cm. This material-independent proximity detection capability enables reliable distance estimation for robotic manipulation tasks, supporting collision avoidance and approach trajectory planning.
    
    \begin{figure}[ht]
        \centering
        \includegraphics[width=1.\columnwidth]{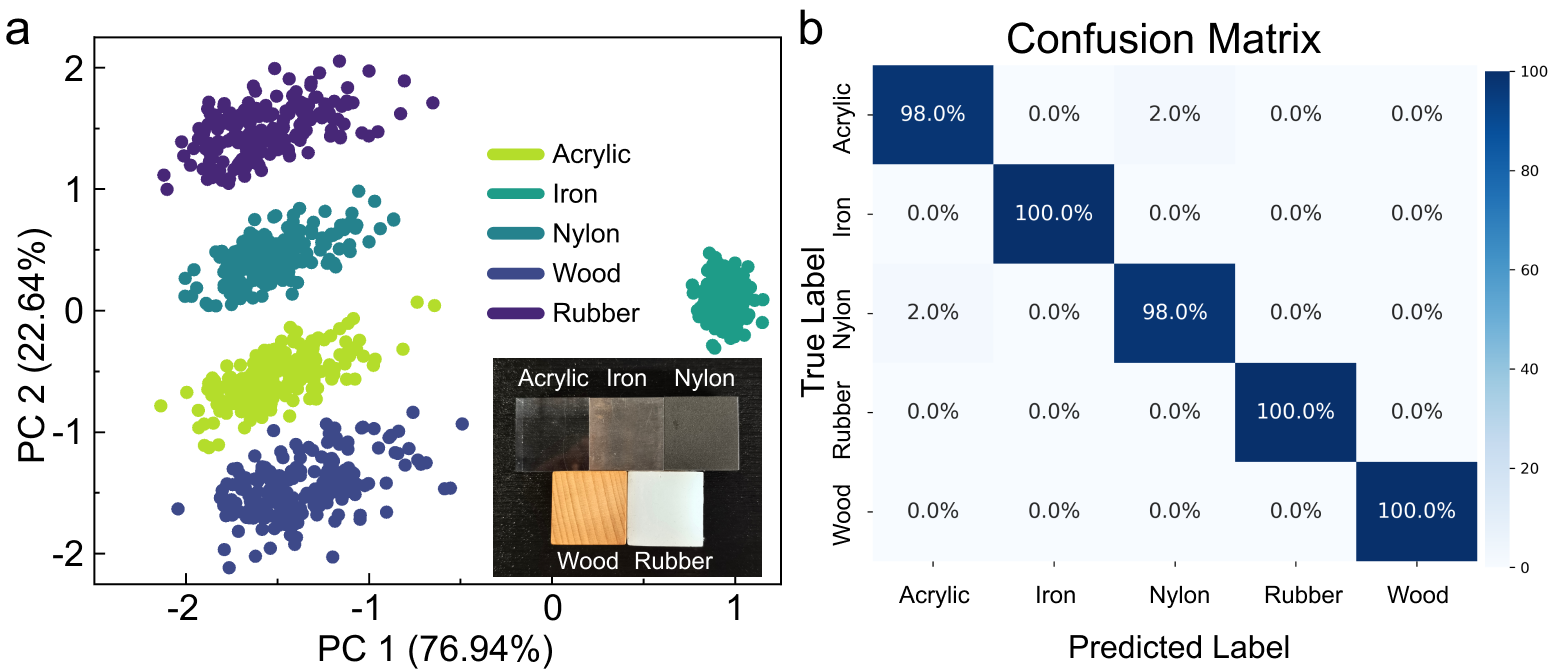} 
        \caption{Material classification experiment. (a) PCA of the spectral features for five materials. The inset shows photographs of the corresponding physical samples. (b) Confusion matrix with 99.20\% average accuracy.}
        \label{fig:mat_class}
    \end{figure}
    
    \vspace{-2mm}

\subsection{Material Classification Experiment} 

 In this experiment, we explore material classification using ultrasound with the aim of capturing additional dimensions of information and overcoming the limitations of traditional tactile sensors, which are confined to detecting only surface characteristics. To minimize interference from properties unrelated to the acoustic characteristics, experiments are conducted using uniform blocks with consistent size and thickness. For each echo signal frame, a Fourier transform is performed to extract spectral features such as spectral contrast, spectral kurtosis, spectral skewness, and spectral entropy. Fig.~\ref{fig:mat_class}(a) demonstrates the clear separability achieved through Principal Component Analysis (PCA), with PC1 and PC2 explaining 76.94\% and 22.6\% of variance respectively across five distinct materials (Acrylic, Iron, Nylon, Rubber, and Wood).

The confusion matrix in Fig.~\ref{fig:mat_class}(b) confirms high classification accuracy: Iron, Rubber, and Wood achieve perfect classification (100\%), while Acrylic (98\%) and Nylon (98\%) show minimal confusion with each other (2\% error rate). These results validate ultrasound's effectiveness for reliable material discrimination, capturing information inaccessible to conventional tactile sensors.

\begin{figure}[ht]
    \centering
    \includegraphics[width=1.\columnwidth]{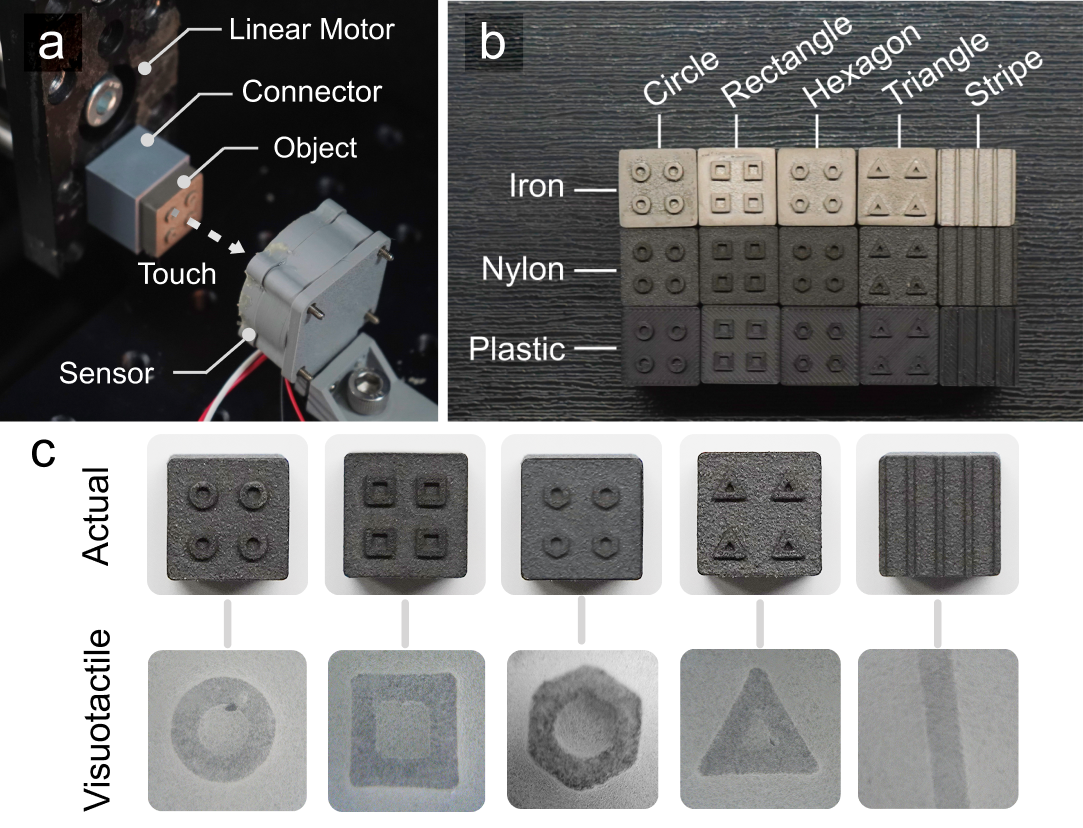} 
    \caption{Dual-modal experiment. (a) Experimental setup. (b) Test objects with patterns across three materials. (c) Actual objects and their corresponding visuotactile results.}
    \label{fig:Demo1-1}
\end{figure}
\vspace{-2mm}

\subsection{Dual-Modal Classification Experiment}

The dual-modal assessment combines tactile and ultrasound sensing to enhance object classification by leveraging both surface pattern recognition and material classification. As illustrated in Fig.~\ref{fig:Demo1-1}(a), the experimental setup comprises a linear motor that facilitates horizontal movement with a cube-shaped test object attached via a mechanical connector. During contact between the sensor and the object, the system concurrently acquires tactile images and ultrasound echo data. To develop the classification models, we collected 200 samples for each of five materials at intervals of 100 ms, capturing both modalities. The dataset was divided using an 8:2 train-test split to ensure robust model evaluation. For the ultrasound data, each frame was transformed via Fourier analysis to extract features as inputs for XGBoost material classification. For the visuotactile data, image frames were preprocessed with data augmentation operations such as resizing, rotation, and noise injection before being input to ResNet model for surface pattern classification. Both training processes were executed on an RTX4070 platform.

\begin{figure}[ht]
    \centering
    \includegraphics[width=0.98\columnwidth]{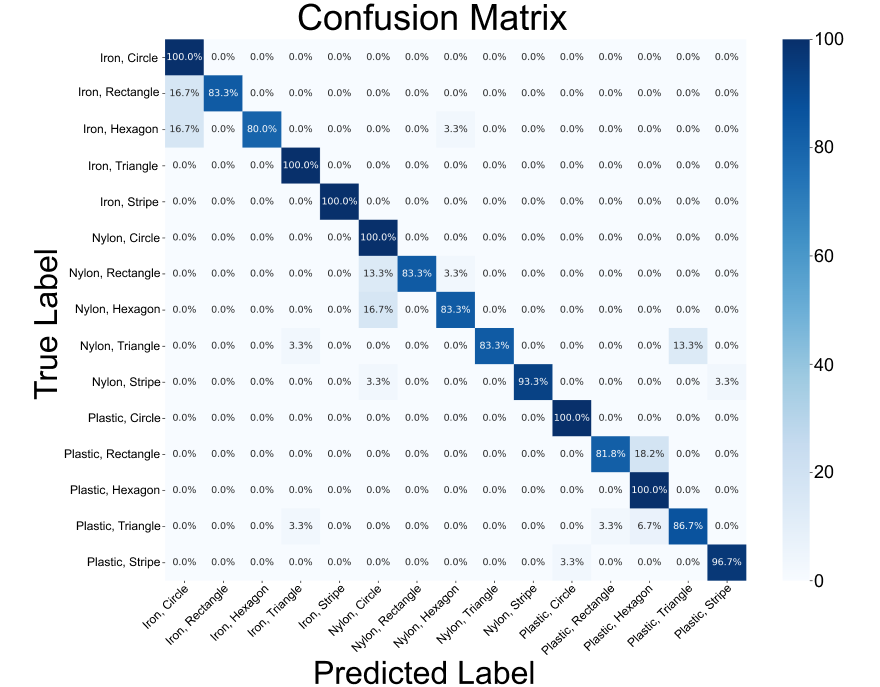}  
    \caption{Confusion matrix for dual-modal experiment with 92.11\% average accuracy.}
    \label{fig:Demo1-3}
\end{figure}

To evaluate the overall system performance, we tested it with various objects as depicted in Fig.~\ref{fig:Demo1-1}(b). These objects feature diverse surface patterns (circle, rectangle, hexagon, triangle, stripe) applied to materials including iron, nylon, and plastic. Fig.~\ref{fig:Demo1-1}(c) presents a comparison between the actual test objects and the corresponding sensor readings, thereby demonstrating the system's capability to accurately capture and classify surface geometry and material characteristics.

The classification results in Fig.~\ref{fig:Demo1-3} demonstrate the effectiveness of this approach in a 15-class classification task, achieving an average accuracy of 92.11\%. The confusion matrix shows strong performance across all materials and shapes, with minor misclassifications occurring primarily between shapes of the same material. This high accuracy confirms that the integration of tactile and ultrasound sensing enables robust discrimination of both geometric features and material properties.

\begin{figure}[ht]
    \centering
    \includegraphics[width=1.\columnwidth]{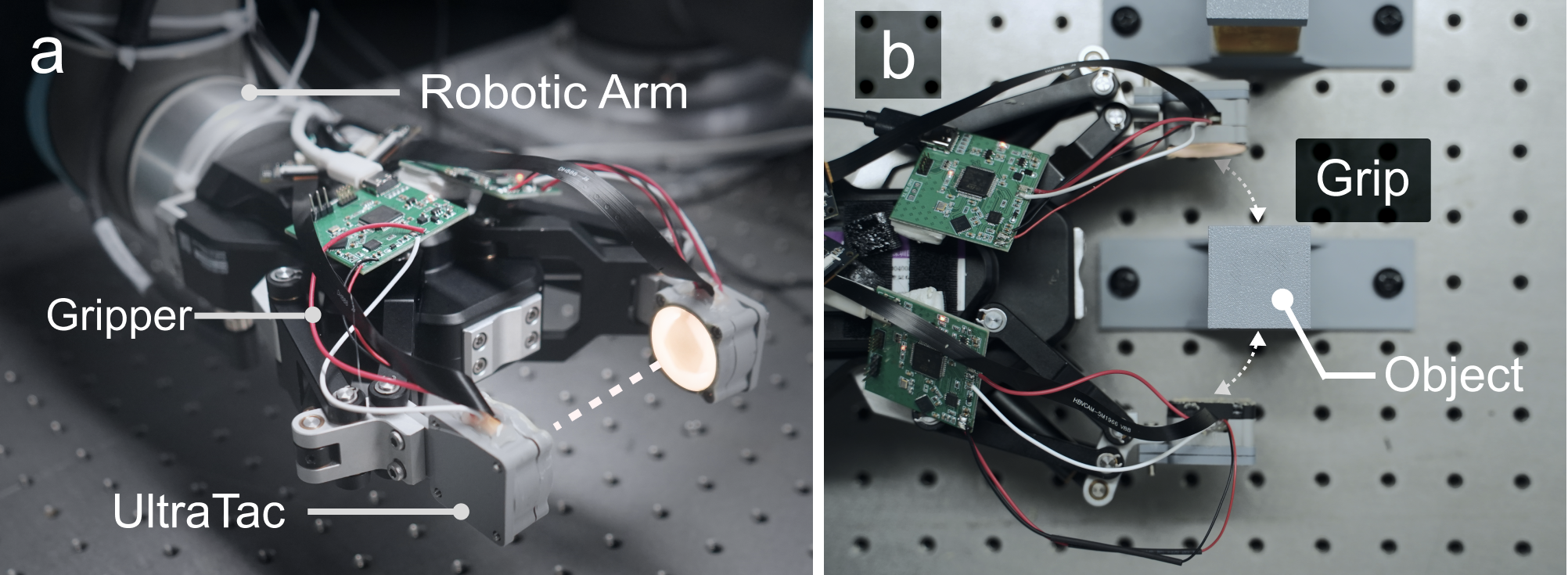} 
    \caption{Internal contents inspection experiment. (a) Robotic gripper setup. (b) Gripper grasping test object.}
    \label{fig:Application1}
\end{figure}
\vspace{-2mm}

\begin{figure*}[ht]
    \centering
    \includegraphics[width=2.\columnwidth]{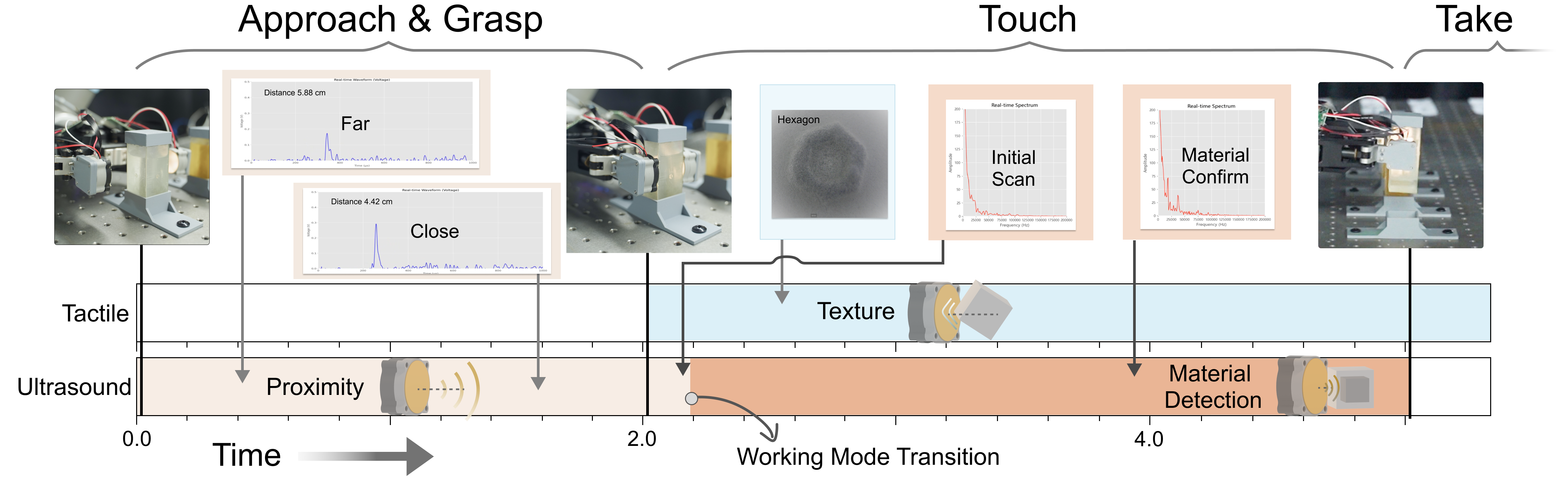} 
    \caption{Timeline of the system operation during internal contents inspection: Approach \& Grasp (ultrasound distance measurement), Touch (tactile texture recognition and ultrasound material detection after working mode transition), and Take (transport execution). }
    \label{fig:Application3}
\end{figure*}

\begin{figure}[ht]
    \centering
    \includegraphics[width=1.\columnwidth]{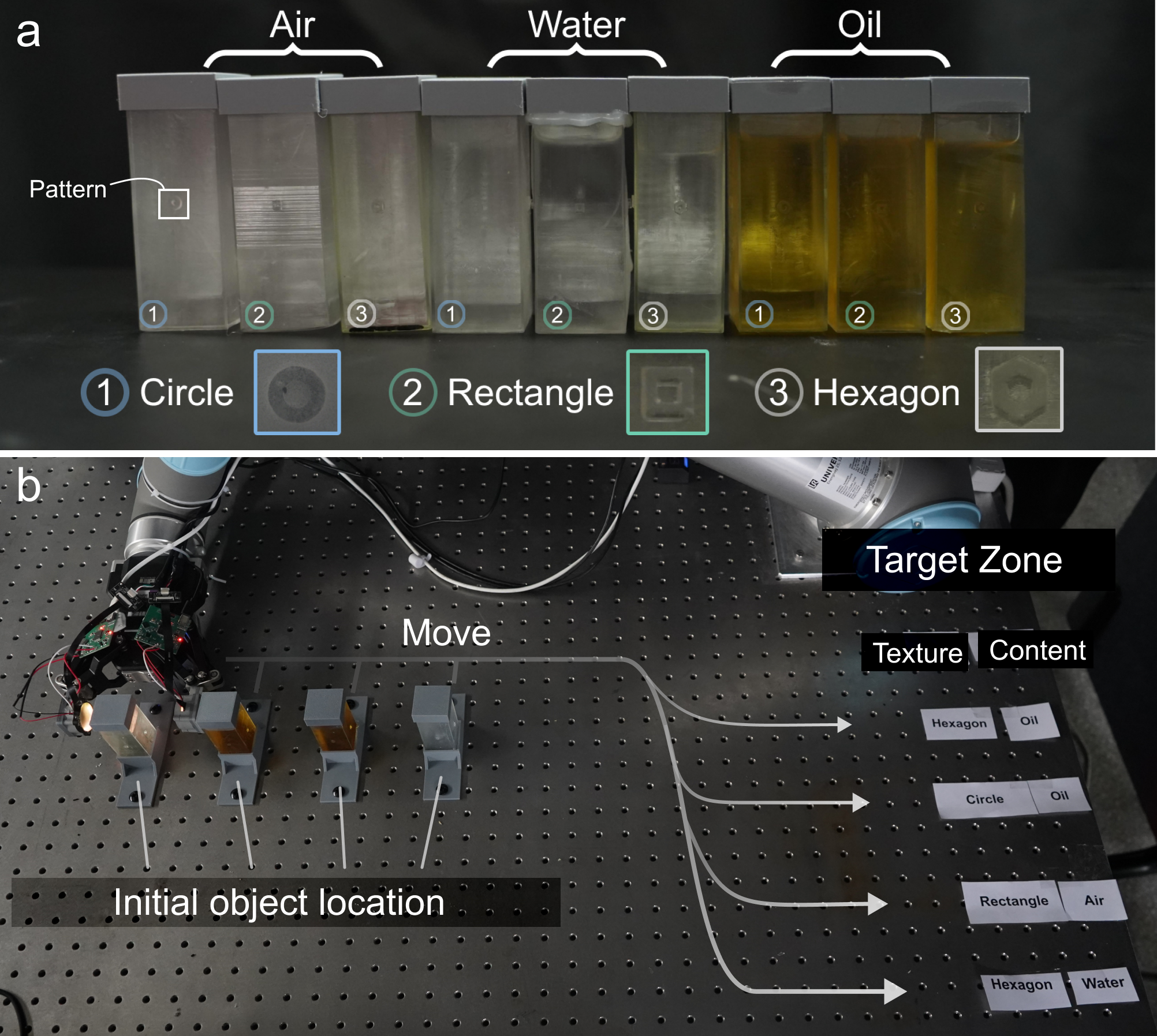} 
    \caption{Test set and workflow. (a) Test containers with varied contents and patterns. (b) Workflow showing transport from initial positions to target zones sorted by texture and content.}
   \label{fig:Application2}
\end{figure}

\subsection{Internal Content Inspection Experiment}

The internal contents inspection experiment demonstrates UltraTac's capability for automated quality control by identifying container surface patterns and internal contents without opening packages. As shown in Fig.~\ref{fig:Application1}(a), two UltraTac sensors are mounted on a robotic gripper attached to a robotic arm for validation. Fig.~\ref{fig:Application1}(b) shows the gripper interacting with test objects during grasping, with sensors acquiring both tactile and ultrasound data.

The inspection process follows a structured timeline as illustrated in Fig.~\ref{fig:Application3} and comprises three distinct phases: approach and grasp, touch, and take. In the approach and grasp phase (0-2 s), the ultrasound modality provides proximity sensing to ensure an optimal grasp position by employing ToF calculations to accurately determine the distance. Upon contact at 2 s, the sensor system undergoes a working mode transition. In the subsequent touch phase, the tactile modality captures the surface textures, as exemplified by the hexagonal pattern, while the ultrasound modality switches to a material detection mode to analyze the internal contents by leveraging the spectral characteristics of the echo signals. Finally, during the take phase, the robotic arm executes the transport operation, delivering the container to its designated location based on both its surface pattern and internal content. 

We created a test set of nine containers featuring various internal contents (air, water, oil) and surface patterns (circle, rectangle, hexagon), as illustrated in Fig.~\ref{fig:Application2}(a). Each content type includes all three surface patterns, which poses challenges for discrimination: identical patterns with different contents (for example, hexagon-patterned containers containing different substances) and identical contents with different patterns (such as various patterns all containing oil). Fig.~\ref{fig:Application2}(b) presents the workflow in which containers are transported from their initial positions to target zones, sorted according to both texture and content type (operation videos and results are listed on the project website). The successful placement of each container underscores the efficacy of the sensor system in concurrently identifying surface features and internal properties, thereby enhancing the overall performance of the robotic application.

\section{LIMITATIONS AND FUTURE WORKS}

Ultrasound sensing complements visuotactile perception by adding proximity detection and internal inspection, but it has limits: echoes from targets closer than 3 cm fall into a blind zone due to pulse duration and receiver recovery, and HGM fillers—whose particles are larger than conventional ones—degrade imaging resolution. Future improvements such as increasing excitation voltage, reducing transducer size, or optimizing transducer placement may alleviate the need for strict acoustic matching while improving overall sensor performance.

\section{CONCLUSIONS}

In this paper, we introduced UltraTac, an integrated ultrasound-augmented visuotactile sensor based on a novel coaxial optoacoustic architecture that combines high-resolution tactile imaging with ultrasound proximity sensing. By aligning a micro-camera and a ring-shaped PZT transducer along a common optical axis and employing innovative acoustic matching techniques within a flexible, optically transparent membrane, UltraTac overcomes the key limitations of conventional visuotactile sensors, which cannot capture material properties or perform proximity sensing. Our dual-pathway signal processing pipeline further enables dynamic switching between proximity detection and material classification modes based on tactile feedback.

Systematic experimental evaluations demonstrate that UltraTac delivers robust performance across multiple sensing tasks. Proximity detection experiments revealed a strong linear correlation (R² = 0.99) between estimated and actual distances, while material classification based on Fourier-derived spectral features achieved 99.2\% accuracy across various materials. Dual-modal object recognition experiments confirmed the sensor’s capability with a 92.11\% accuracy in a 15-class task, and integration into a robotic gripper showcased its practical utility for automated quality control by successfully identifying both container surface patterns and internal contents without opening packages. Future work will focus on further miniaturization, refinement of signal processing algorithms, and integration of additional sensing modalities to enhance tactile perception in unstructured, real-world environments, paving the way for safer, more efficient, and intelligent human–machine interaction systems.



\end{document}